\documentclass[11pt,letterpaper]{article}
\usepackage{emnlp2016}
\usepackage{times}
\usepackage{latexsym}
\usepackage{booktabs}
\usepackage{amsmath, amssymb}
\usepackage[pdftex]{graphicx}
\usepackage{layout}
\usepackage{url}

\emnlpfinalcopy

\title{Sparse Named Entity Classification using Factorization Machines}

\author{Ai Hirata \and Mamoru Komachi\\
	    Tokyo Metropolitan University\\
	    6-6 Asahigaoka, Hino, Tokyo 191-0065, Japan\\
	    {\tt hirata-ai@ed.tmu.ac.jp \and komachi@tmu.ac.jp}}

\date{}

\begin{document}

\setlength{\abovedisplayskip}{4pt} % 上部のマージン
\setlength{\belowdisplayskip}{4pt} % 下部のマージン

\maketitle

\begin{abstract}
Named entity classification is the task of classifying text-based elements
into various categories , including places, names, dates, times, and
monetary values. A bottleneck in named entity classification, however,
is the data problem of sparseness, because new named entities continually
%固有表現認識は自然言語処理の重要なタスクである．
%
emerge, making it rather difficult to maintain a dictionary for named entity
classification.
%しかし，日々書かれる文章には学習コーパスに出現しない固有表現が増えていくた
%め，そのような固有表現にも正しい固有表現タグを割り当てる必要がある．
%
Thus, in this paper, we address the problem of named entity classification
%そこで，この研究では学習データに出現しない固有表現に対して固有表現分類を行う．
%
using matrix factorization to overcome the problem of feature sparsity.
%学習データに出現しない固有表現を分類するためには，前後の文脈や文字種素性から
%
Experimental results show that our proposed model, with fewer features and
a smaller size, achieves competitive accuracy to state-of-the-art models.
%実験の結果から，Factorization Machines を用いることで先行研究よりも高い精度が
%得られた．
\end{abstract}

\section{Introduction}
To date, standard approaches to named entity classification rely on supervised
models, that typically require a large-scale annotated corpus and a
wide-coverage dictionary.
%これまで，教師あり学習における固有表現認識の研究が行われてきたが，学習コーパ
%スとして大量のタグ付きデータが必要である．
%
However, since new named entities arise regularly, it becomes increasingly
difficult to maintain an up-to-date dictionary and/or adapt a named entity
classifier to a new domain;
%しかし，固有表現認識の課題として，日々新しい固有表現が登場するので，学習
%コーパスに出現しなかった固有表現に対しても正しい予測をする必要がある．
%
for example, sequence labeling techniques that use feature templates
\cite{finkel:2005:acl,sarawagi:2004:nips} are not robust for unknown named
entities because their feature space is very sparse \cite{primadhanty:2015:acl}.
This problem worsens when we attempt to use a combination of features
for sparse named entity classification.
%固有表現認識では素性テンプレートを用いた系列ラベリングによるアプローチ
%\cite{knp}が主流である．素性テンプレートを用いれば，___固有表現の前後の単語
%___を素性の組み合わせで表すことができるが，素性空間が膨大でとても疎になってし
%まい，未知の固有表現に対して頑健性が低いという問題がある．

Therefore, in this paper, we propose the use of matrix factorization for named
entity classification to consider the relationships between sparse features.
Through our experiments, we achieved competitive accuracy to models developed
in previous works in terms of using fewer features and compactness using
factorization machines \cite{rendle:2010:icdm}.
%
%そこで，この問題を解決するために，本研究ではFactorization Machines を用いて素
%性の間の関係性を考慮して固有表現認識を行うことを提案する．本研究では学習
%コーパスに出現しない固有表現に対して正しい予測をすることで実験，比較を行い，
%Factorization Machines を用いることでスパースな素性と少ない次元数で先行研究と
%同程度の精度を達成した．
%
The main contributions of this paper are as follows:

\begin{itemize}
\item We address the data sparseness problem in unknown named entity
classification using factorization machines.
\item We demonstrate that factorization machines achieve state-of-the-art
performance in sparse named entity classification task using a reduced
feature set and a compact model.
\end{itemize}

\section{Related Work}

A standard approach to named entity classification is to formulate
a task as a sequence labeling problem and use a supervised method,
such as conditional random fields
\cite{lafferty:2001:icml,finkel:2005:acl,sarawagi:2004:nips}.
These studies heavily rely on feature templates for learning combinations of
features; however, since combinations of features in conventional supervised
learning are treated independently, this approach is not robust for named
entities that do not appear in the training data.
% 一般的な日本語固有表現認識の設定ではConditional Random
% Fields（CRF）\cite{crf_ner}やSupport Vector Machines（SVM）\cite{svm_ner}を
% 用いた手法がよく使用されている．これらの研究では素性テンプレートによる組み合
% わせ素性の展開と学習が行われているが，組み合わせ素性同士は独立して扱うため，
% 学習データで出現しなかった固有表現に対して頑健ではない．

%In study of unknown named entity recognition from few seed,
%\newcite{collins:1999:emnlp} learn using co-training from data that do not
%tag. These semi-supervised method can treat combination feature to use
%expressly develop, on the other hand these are problem that can not
%regularization adequately.
%少ないシードから未知の固有表現を当てるという観点での研究で
%は，\newcite{collins:1999:emnlp}がタグのついていないデータからco-training を
%行うことで学習を行っている．これらの半教師あり手法は，組み合わせ素性を陽に展
%開することで扱うことが可能であるが，適切な正則化ができないという問題点があ
%る．

To address the task of unknown named entity classification,
\newcite{primadhanty:2015:acl} explored the use of sparse combinatorial
features.  They proposed a log-bilinear model that defines a score function
considering interactions between features; the score function is regularized
via a nuclear norm on a feature weight matrix. Further, heir method employs
singular value decomposition (SVD)-based regularization to handle the
combination of features. They reported that their regularization achieved
higher accuracy than L1 and L2 regularization, frequently used in natural
language processing \cite{okanohara:2009:naacl}.

%\newcite{primadhanty:2015:acl}は学習コーパスに出現しなかった未知の固有表現を
%認識するタスクに取り組んだ．
%この研究では対数線形モデルを改良しており，前後の文脈，文字種情報などの素性か
%らスコア関数を定義し，テンソルとして構成された重みを行列化し特異値分解（SVD）
%を用いて核ノルムを計算し，正則化項として用いることでスパースな素性の組み合わ
%せも考慮し，L1，L2 正則化よりも高い精度が得られることを報告している．

However, nuclear norm regularization (i.e., SVD-based regularization) is not
necessarily the best way to incorporate interactions between features, because
it does not directly optimize classification accuracy. Therefore, our proposed
method treats sparse features using matrix factorization from a different
perspective: we decompose a feature weight matrix using factorization machines
as to directly optimize classification accuracy using a large margin method
similar to support vector machines (SVMs) and passive-agressive algorithms
\cite{vapnik:1995,crammer:2006:jmlr}.

%この手法は組み合わせ素性を扱うために特化した正則化を行っているが，特異値分解
%が最もよい正則化であるとは限らない．提案手法も行列分解を行うことでスパースな
%素性を扱うが，分類精度を直接向上させるように分解し，かつ尤度最大化ではなく
%マージン最大化によって学習を行う点が異なる．

\section{Factorization Machines}

In this paper, we propose the use of factorization machines
\cite{rendle:2010:icdm} for unknown named entity classification.  Using this
approach, we can employ the same objective function as SVMs and yet performs
matrix factorization to handle sparse combinatorial features.
Matrix factorization yields better generalizations over a sparse
feature matrix \cite{madhyastha:2014:coling}.

%Factorization Machines \cite{factorization}とはSupport Vector Machines (SVM)
%と行列分解 (Matrix Factorization)を組み合わせたモデルであり，スパースなデータ
%に対応することができるモデルである．

Factorization machines with interaction degree $d = 2$ use
the following equation for prediction:
%
%相互作用の次元を$d = 2$ とした場合のFactorization Machines の予測式は
%
\begin{equation}
\label{eq:fm_equation}
\hat{y}({\bf x}) := w_0 + \sum_{i=1}^n w_ix_i + \sum_{i=1}^n \sum_{j=i+1}^n \langle v_i, v_j \rangle x_i x_j
\end{equation}
\noindent Here, ${\bf x}$ is an instance,
$x_i$ represents the $i$-th dimension of the feature $x$,
$n$ is the number of features,
$w \in \mathbb{R}^n$ is a weight vector, and
$w_0 \in \mathbb{R}$ is a bias term.
%
%\noindent で表される．式\ref{eq:fm_equation}の$n$は素性の次元数，$x_i$は素性
%$x$ の$i$ 番目の次元を表している．
%
%線形モデルと同様に，$w \in \mathbb{R}^n$ で表される$w$ は式
%\ref{eq:fm_equation}での重みベクトルであり，第1項目の$w_0 \in \mathbb{R}$ は
%バイアス項，第2項目の$w_i$ は$x_i$ の重みを表している．
%
Factorization machines incorporate interactions between variables $v_i,
v_j$ as the third term of Equation (\ref{eq:fm_equation}). Here,
$\langle \cdot,\cdot\rangle$ is the inner product of two vectors of size $k$,
i.e.,
%
%そして第3項目は$v_i, v_j$ の変数の交互作用をモデルに組み込んでいる．$\langle
%\cdot,\cdot\rangle$ はサイズ$k$ の2つのベクトルの内積であり，
%
\begin{equation}
\label{eq:inner_product}
\langle v_i, v_j \rangle := \sum_{f=1}^k v_{i,f} \cdot v_{j,f}
\end{equation}
\noindent where $v_i$ is the $i$-th element of matrix ${\bf V}
\in \mathbb{R}^{n\times k} $ and $k$ is a hyperparameter representing
the dimension of matrix decomposition.
%
%\noindent で表される．$v_i$ は${\bf V} \in \mathbb{R}^{n\times k} $ で表され
%る行列の$i$ 番目の要素であり，$k$ は行列分解したあとの次元数を表すハイパーパ
%ラメータである．
%
To consider interactions between features, we only need to calculate
the inner product of a decomposed matrix $n \times k$ times.
Therefore, we do not incur high computational costs even though the
number of interacting features is large.

%行列分解された{\bf V}を用いて交互作用の重みを内積で計算することで，$n \times
%k$ のサイズの計算をするだけでよく，考慮する交互作用の数が増えても計算量が増え
%ない利点がある．

Note that even though the polynomial kernel of SVMs take combinations of
features into account, it treats them independently. Conversely, factorization
machines take advantage of interactions between features using a
low-dimensional feature matrix via matrix factorization.
%
%また，多項式カーネルのSVM では素性間の相互関係が独立であるが，Factorization
%Machinesではテンソル分解を用いる手法と同様，行列分解された低次元の行列を用い
%ることで素性間の相互関係を学習できるという違いがある．
%
Because factorization machines can learn combinations of infrequent features
thanks to matrix factorization, we expect that factorization machines will
correctly classify named entities that seldom appear in the training corpus.
%Factorization Machines では学習コーパスに出現しない素性の組み合わせも考慮でき
%るので，今回のタスクの学習コーパスに出現しなかった未知の固有表現をうまく認識
%できると期待される．

In a binary classification task, factorization machines use hinge loss
to optimize parameters. Here, parameter learning can be accomplished via
Markov chain Monte Carlo or stochastic gradient descent.

%今回の実験では2値分類を行うが，Factorization Machines で2値分類を行うときはヒ
%ンジロスを計算し，最適化を行う． 
%最適化にはMarkov Chain Monte Carlo (MCMC) やStochastic Gradient Descent (SGD)
%などで推定を行う．

\section{Experiments}

\begin{table}
\centering
  \scalebox{0.9}{
  \begin{tabular}{crrrr}
  \toprule
  & \multicolumn{2}{c}{training} & development & \multicolumn{1}{c}{test} \\
  \midrule
  \textsc{per}  &  6,516 & (3,489) & 1,040 (762) & 1,342 (925) \\
  \textsc{loc}  &  6,159 &   (987) &   176 (128) &   246 (160) \\
  \textsc{org}  &  5,721 & (2,149) &   400 (273) &   638 (358) \\
  \textsc{misc} &  3,205 &   (760) &   177 (142) &   213 (152) \\
  \textsc{o}    & 36,673 & (5,821) &   951 (671) &   995 (675) \\
  \bottomrule  
  \end{tabular}
  }
    \caption{Number of candidates (i.e., tokens) in the dataset obtained from
    \protect\newcite{primadhanty:2015:acl}, with the number of unique
    candidates (i.e., types) shown in parentheses.}
  \label{tab:data}
\end{table}

\begin{table}[t]
\centering
  \scalebox{0.80}{
  \begin{tabular}{cccc}
  \toprule
                   & P & R & F1  \\
  \midrule
  log-linear model & 49.75 & 44.50 & 46.75 \\
  SVM (polynomial kernel) & 53.75 & 50.67 & 51.94 \\ 
  \shortstack{log-bilinear model \\ \cite{primadhanty:2015:acl}} &
    \raisebox{0.5em}{{\bf 62.03}} & \raisebox{0.5em}{53.92} &
    \raisebox{0.5em}{55.88} \\
  factorization machines & 60.93 & {\bf 55.10} & {\bf 57.27}  \\
  \bottomrule
  \end{tabular}
  }
  \caption{Results of unknown named entity classification.}
  \label{tab:result}
\end{table}

\begin{table*}[t]
\centering
  %\small
  \begin{tabular}{|p{\textwidth}|} \hline
  \textbf{context features}: Right and left contexts of the 
      candidate in a sentence (do not take the order into account).\\
  \textbf{cap=1, cap=0}: Whether the first letter of the  candidate is uppercase, or not. \\
  \textbf{all-low=1, all-low=0}: Whether all letters of the 
      candidate are lowercase, or not. \\
  \textbf{all-cap1=1, all-cap1=0}: Whether all letters of the
      candidate are uppercase, or not. \\
  \textbf{all-cap2=1, all-cap2=0}: Whether all letters of the
      candidate are uppercase and periods, or not. \\
  \textbf{num-tokens=1, num-tokens=2, num-tokens$>$2}: Whether the
      candidate consists of 1, 2, or more tokens. \\
  \textbf{dummy}: Dummy feature to capture context features. \\
  \hline
  \end{tabular}
    \caption{Features used in our experiment; note that this is a subset of
    features used in \protect\newcite{primadhanty:2015:acl}'s experiment.}
  \label{tab:feature}
\end{table*}

As described above, we aim to classify named entities that rarely appear in a
given training corpus.
%今回のタスクとして，学習コーパスに出現しない固有表現の認識を行う．
%
We compared factorization machines with a log-linear model, a
polynomial-kernel SVM, and a state-of-the-art log-bilinear model using nuclear
norm for regularization \cite{primadhanty:2015:acl}.
%比較手法として，対数線形モデル，\newcite{primadhanty:2015:acl} の素性をテンソ
%ルとして用いた対数双線形モデル（核ノルム正則化）と多項式カーネルを用いたSVM，
%Factorization Machines を比較する．

\subsection{Settings}

\paragraph{Data.} We used the dataset provided by
\newcite{primadhanty:2015:acl}; this dataset was created for evaluating
unknown named entity classification and 
%今回使用するデータは \newcite{primadhanty:2015:acl} が作成したデータを使う．
%
is based on the CoNLL-2003 English dataset, which omits named entity
candidates that appear in the training data from the development and test
data.
%Note that we follow the same training/development/test split of the
%riginal set.
%\newcite{primadhanty:2015:acl}のデータはCoNLL-2003 をもとにして，テストデータ
%と開発データから学習データに出現した固有表現の候補を取り除いたものである．

Table \ref{tab:data} shows the number of tokens and types in the given
dataset.  This dataset contains five tags: person (\textsc{per}), location
(\textsc{loc}), organization (\textsc{org}), miscellaneous (\textsc{misc}),
and non-entities (\textsc{o}).
%このデータには，人名（PER），地名（LOC），組織名（ORG），その他の固有表現
%（MISC），固有表現以外（O）のタグが固有表現の候補となるフレーズに割り当てられ
%ている．学習データとテストデータ，開発データのタグの割合を表\ref{tab:data}に示す．
%

\paragraph{Features.}
We used a subset of features from experiments performed by
\newcite{primadhanty:2015:acl}.
Table \ref{tab:feature} summarizes the features used in our experiment,
including context and entity features.
% 今回行う実験の素性は \newcite{primadhanty:2015:acl} の実験と同様の素性を用いる．
%大きく分けて文脈素性と12 種類の文字種素性があり，表\ref{feature}の通りである．

%\begin{table*}
%\begin{center}
% \caption{Features in this experiment}
% \label{feature}
% \small
% \begin{tabular}{|p{16.5cm}|} \hline
%  \textbf{文脈素性}：識別する固有表現候補が含まれる文の固有表現の順番を考慮しない右文脈，左文脈（単語が出現したかしないかのバイナリ素性として用いる） \\
%  \textbf{cap=1, cap=0}：固有表現候補の単語の最初の文字が大文字かどうか \\
%  \textbf{all-low=1, all-low=0}：固有表現候補の単語が全て小文字かどうか \\
%  \textbf{all-cap1=1, all-cap1=0}：固有表現候補の単語が全て大文字かどうか \\
%  \textbf{all-cap2=1, all-cap2=0}：固有表現候補の単語が全て大文字でかつ，ピリオドがあるかどうか \\
%  \textbf{num-tokens=1, num-tokens=2, num-tokens$>$2}：固有表現候補の単語が1 単語で構成されているか，または2 単語か，それ以上か \\
%  \textbf{dummy}：\newcite{primadhanty:2015:acl} の実験で必要な文脈や文字種素性がない場合のダミー素性 \\ \hline
%  \end{tabular}
%\end{center}
%\end{table*}
 
\paragraph{Tools.}
In terms of tools, we used scikit-learn 0.17 to implement a log-linear model
and polynomial kernel in an SVM. Further, we employed libFM
1.4.2\footnote{\url{http://www.libfm.org/}}
\cite{rendle:2012:tist} to build a named entity classifier using
factorization machines.
%対数線形モデル，多項式カーネルを用いたSVM はscikit-learn （Version 0.17）を用
%いて実験を行う．Factorization Machines のツールとしてlibFM（Version
%1.4.2）\cite{rendle:2012:tist} を用いる．

In the interaction of both the SVM and the factorization machine, we fixed the
degree of the polynomial kernel to $d=2$. We also tuned other parameters such
as learning methods, learning rate and regularization methods based on
development data.
%Number of dimensions $k$ of matrix decomposition and method of
%parameter estimation are tuned using development data.
Further, we used a one-versus-all strategy to build a multiclass classifier.
%多項式カーネルSVM，Factorization Machines は開発データでパラメータチューニン
%グを行い，one-vs-all 法を用いて多値分類を行う．

%また，Factorization Machines のパラメータは相互作用の次元は$d=2$で固定し，行
%列分解したあとの次元数である$k$ や，パラメータ推定の手法などは開発データでパ
%ラメータチューニングを行う

\paragraph{Evaluation metrics.}
For our evaluation, we used precision, recall, and F1-score.
The scores were calculated on all tags except for non-entities
(\textsc{o}).
%今回，実験の評価指標としてprecision とrecall とF1 スコアを用いて比較を行う．
%固有表現以外（O）を除いた4種類のタグを用いて評価する．

\subsection{Results}

\begin{figure}[t]
 \centering
  \includegraphics[width=7.3cm]{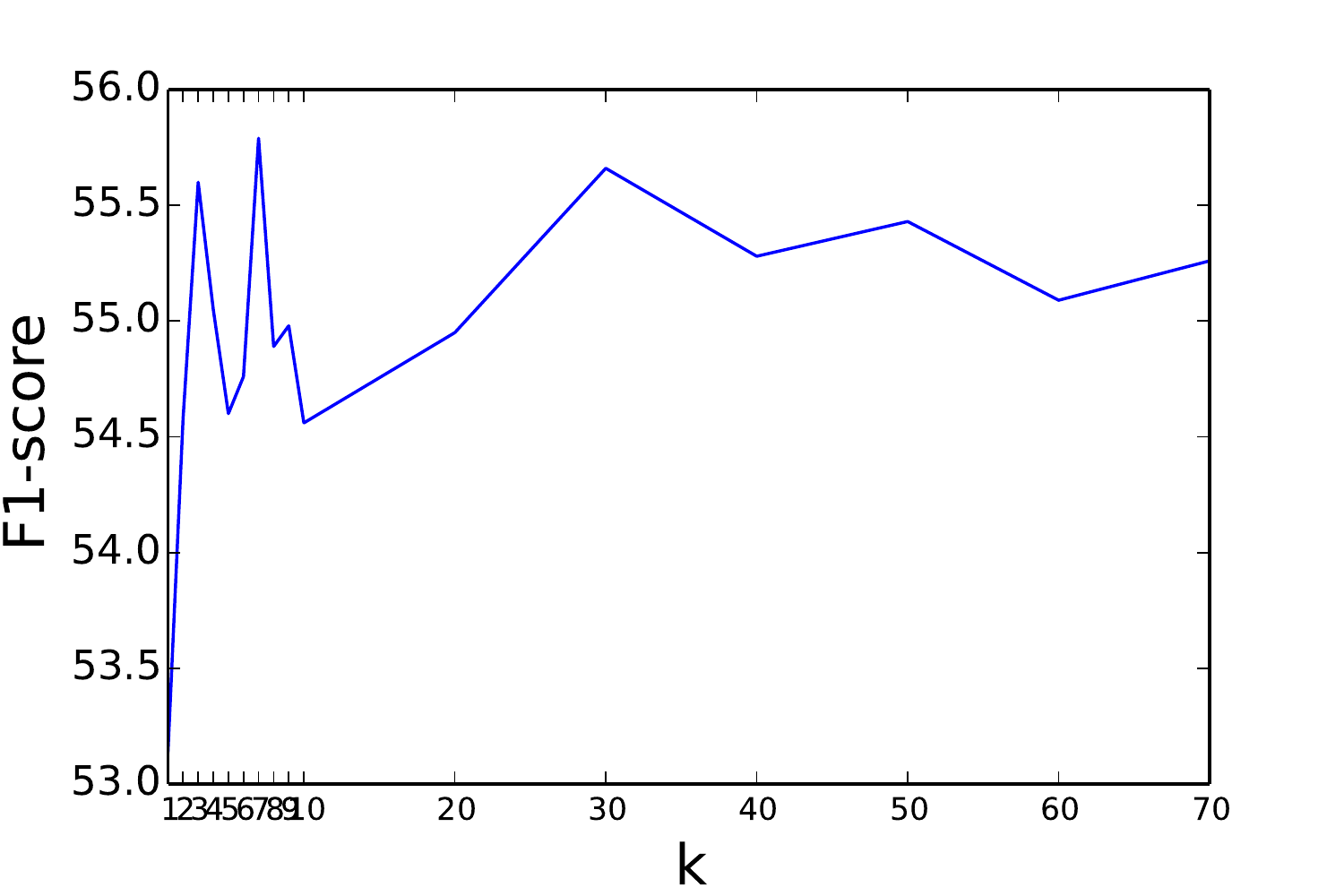}
%  \caption{Factorization Machines における次元数$k$ を変化させた時の精度}
  \caption{F1-score on development data as varying dimension $k$
  varies using factorization machines.}
  \label{fig:dimension}
\end{figure}

\begin{table*}[t]
  \centering
  \scalebox{0.75}{
   \begin{tabular}{c|ccc|ccc|ccc|ccc}
   \toprule
   & \multicolumn{3}{|c}{\textsc{person}} &
   \multicolumn{3}{|c}{\textsc{location}} &
   \multicolumn{3}{|c}{\textsc{organization}} &
   \multicolumn{3}{|c}{\textsc{misc}} \\
   & P & R & F1 & P & R & F1 & P & R & F1 & P & R & F1 \\
   \midrule
   %%%%%%%%%%%%%%%%%%%%%%%%%%%
   SVM (polynomial kernel) & {\bf 86.45} & 72.28 & 78.73 & 31.35 & 38.62 & 34.61 & 62.54 & {\bf 59.40} & 60.93 & 34.67 & 32.39 & 33.50 \\
   %%%%%%%%%%%%%%%%%%%%%%%%%%%
   \shortstack{log-bilinear model \\ \cite{primadhanty:2015:acl}} &
   \raisebox{0.5em}{73.83} & \raisebox{0.5em}{{\bf 90.84}} &
   \raisebox{0.5em}{81.46} & \raisebox{0.5em}{{\bf 64.96}} &
   \raisebox{0.5em}{36.18} & \raisebox{0.5em}{{\bf 46.48}} &
   \raisebox{0.5em}{{\bf 72.11}} & \raisebox{0.5em}{44.98} &
   \raisebox{0.5em}{55.41} & \raisebox{0.5em}{37.20} & \raisebox{0.5em}{{\bf
   43.66}} & \raisebox{0.5em}{40.17} \\
   %%%%%%%%%%%%%%%%%%%%%%%%%%%
   factorization machines & 84.36 & 80.40 & {\bf 82.33} & 39.49 & {\bf 50.41} & 44.29 & 70.88 & 55.33 & {\bf 62.15} & {\bf 48.99} & 34.27 & {\bf 40.33} \\
   \bottomrule
   \end{tabular}
   }
  \caption{A breakdown of the results of unknown named entity classification per tag.}
  \label{tab:result:tag}
\end{table*}

%%あとで画像いれる
\begin{figure*}
 \small
 \begin{tabular}{cc}
 
 \begin{minipage}{.5\textwidth}
 \centering
  \includegraphics[width=7.3cm]{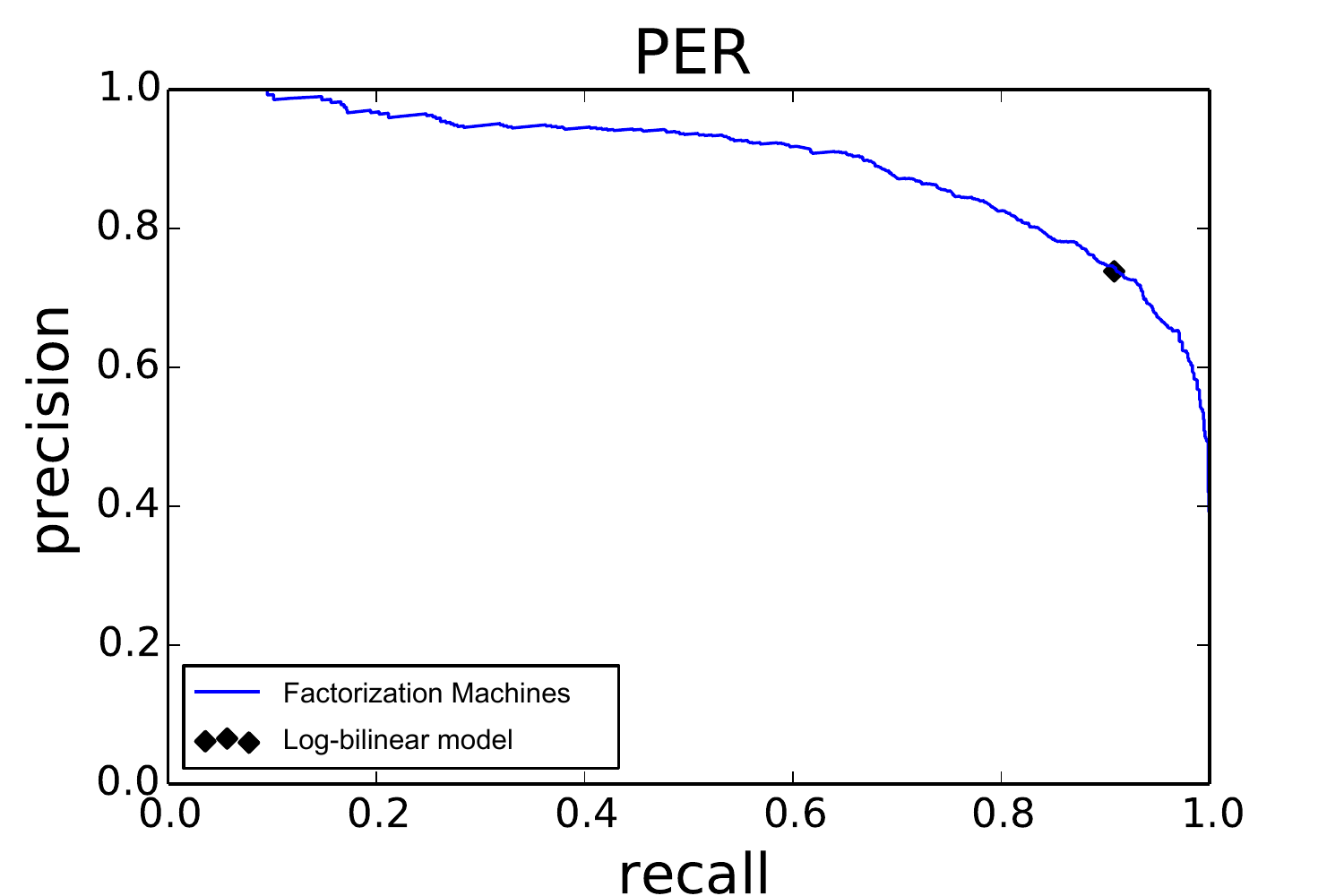}
%  \caption{PERSON タグにおけるprecision-recall 曲線}
%  \label{result_person}
 \end{minipage}

 \begin{minipage}{.5\textwidth}
 \centering
  \includegraphics[width=7.3cm]{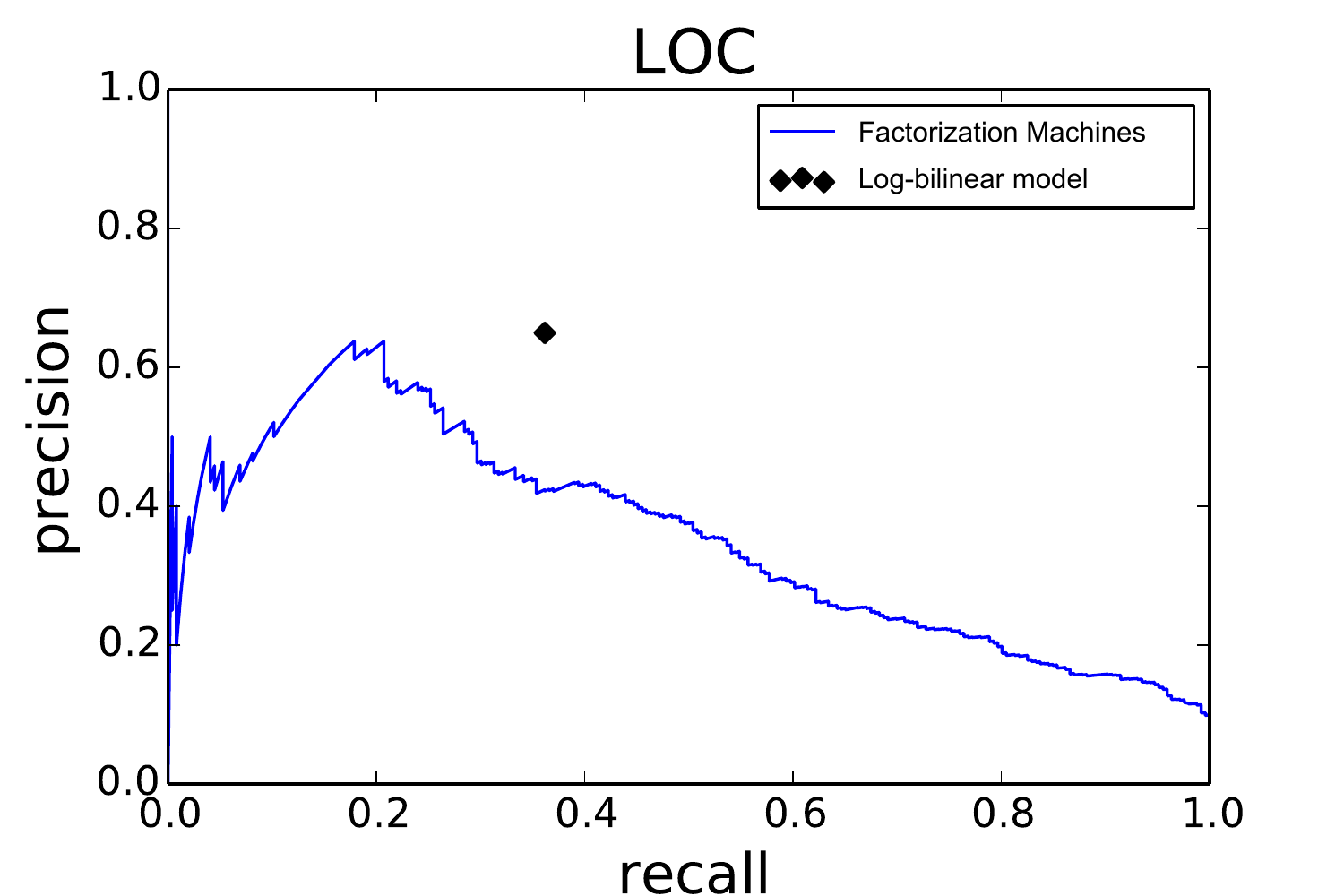}
%  \caption{LOCATION タグにおけるprecision-recall 曲線}
%  \label{result_location}
  \end{minipage}
  
 \end{tabular}

 \small
 \begin{tabular}{cc}
 
 \begin{minipage}{.5\textwidth}
 \centering
  \includegraphics[width=7.3cm]{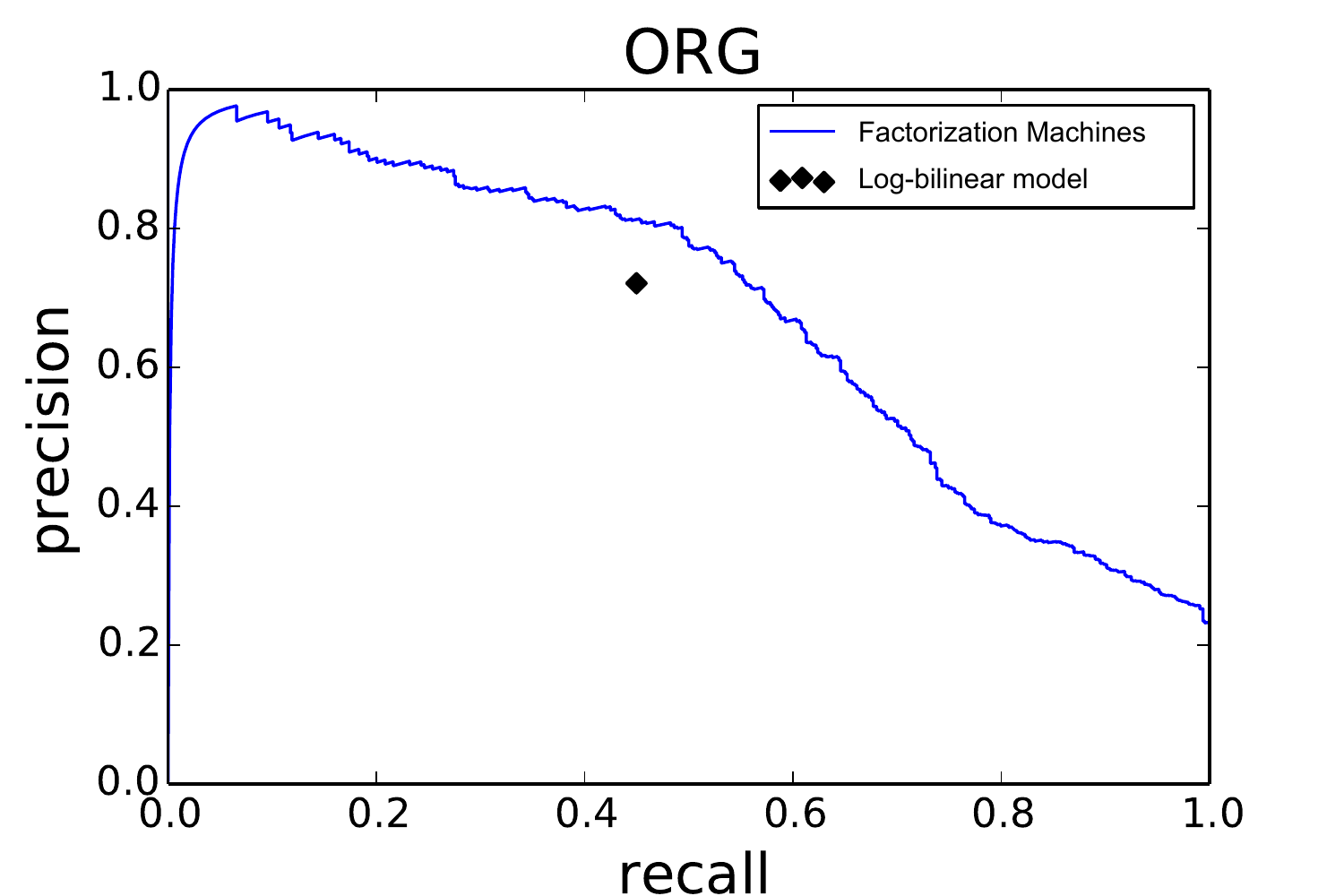}
%  \caption{ORGANIZATION タグにおけるprecision-recall 曲線}
%  \label{result_org}
  \end{minipage}

  \begin{minipage}{.5\textwidth}
  \centering
   \includegraphics[width=7.3cm]{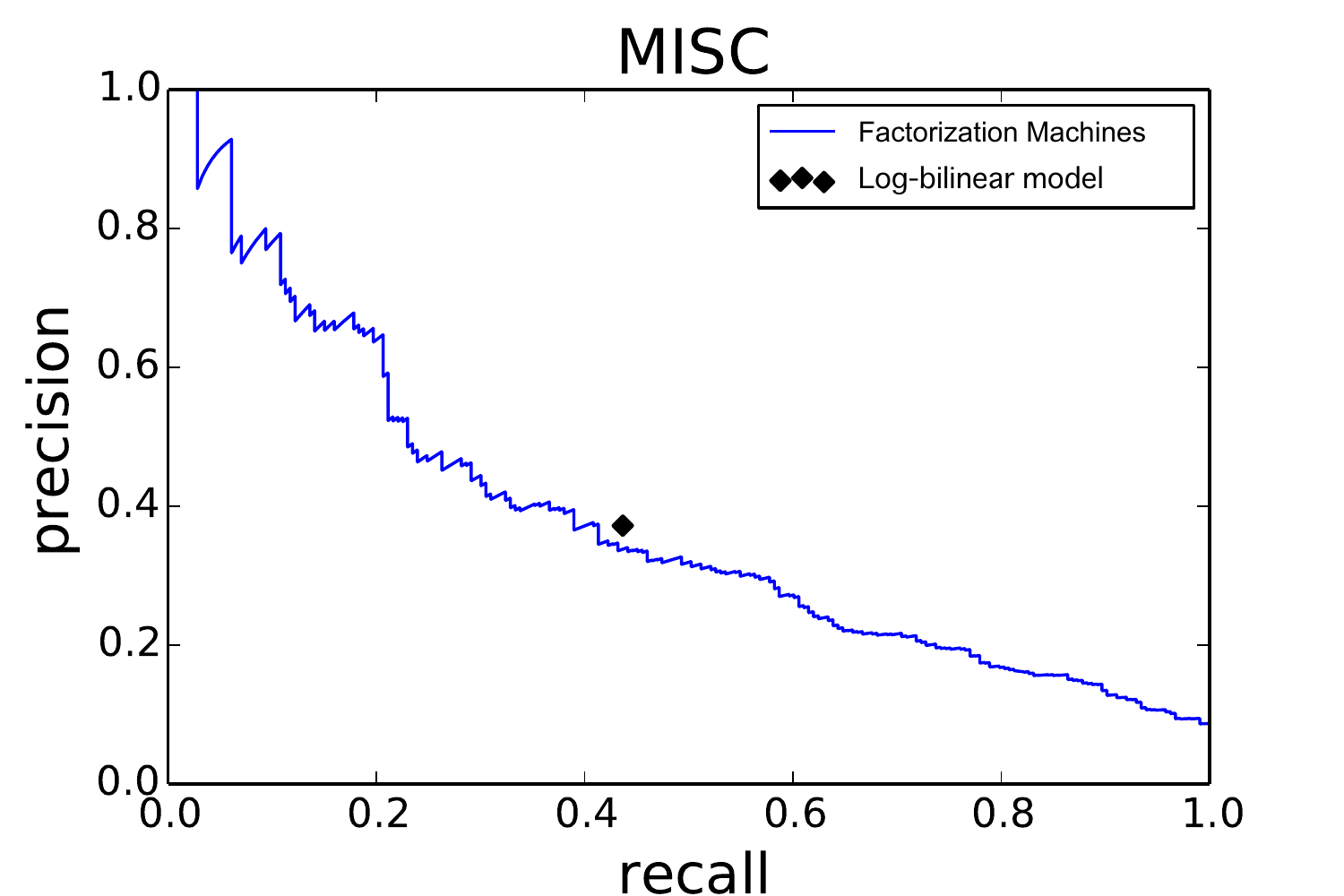}
%   \caption{MISC タグにおけるprecision-recall 曲線}
%  \label{result_misc}
  \end{minipage}
  
 \end{tabular}

\caption{Precision-recall curves of unknown named entity classification for each tag.}
\label{fig:precision_recall_curve}
\end{figure*}

Table \ref{tab:result} presents results of our experiments.
Note that \newcite{primadhanty:2015:acl} used additional features such as
Brown clustering and parts-of-speech (POS) features, which we did not
use.
%表\ref{tab:result}が \newcite{primadhanty:2015:acl} のデータで実験した結果で
%ある．\footnote{\newcite{primadhanty:2015:acl} の実験ではクラスタリング素性や
%POS タグなどの素性も含まれているが，本研究の実験では使用していない．}
%
Table \ref{tab:result:tag} and Figure \ref{fig:precision_recall_curve}
show the performance and precision-recall curves of named entity
classification for each tag, respectively.
%また，表\ref{tab:result:tag}が固有表現タグごとの結果である．図
%\ref{fig:precision_recall_curve}がそれぞれのタグごとのprecision-recall 曲線で
%ある．

We observed here that, aside from \textsc{loc}, we obtained competitive
results to the state-of-the-art named entity classifier proposed by
\newcite{primadhanty:2015:acl} with fewer features.
Overall, the micro-averaged F1 score improved by 1.4 points.
%今回の実験の結果より，Factorization Machines と先行研究の
%\newcite{primadhanty:2015:acl} の手法とで同程度の結果が得られることが分
%かった．precision が1 ポイント低下するものの，recallが1 ポイント改善し，F1 で
%は1 ポイントの改善となっている．

From these results, we conclude that unknown named entity classification can
be successfully achieved by taking combinatorial features into account using
factorization machines.
%この結果よりFactories Machines を用いてスパースなデータでも交互作用を考慮し，
%未知の固有表現をうまく認識できることが分かった．

\section{Discussion}

Experimental results show that performance on \textsc{org} was improved. For
example, the term ``Vice-President'' appears in both contexts of \textsc{org}
and \textsc{o}, and our method correctly handled this sparse combination of
context and entity features.

%実験の結果からタグ``ORGANIZATION''の結果が特に向上しているが，たとえば
%``ORGANIZATION''タグとと``OTHER''タグが付いた事例両方に``Vice-President''とい
%う単語が前後の文脈に出現している．このとき，``Vice-President''以外の前後の文
%脈も似ているが，提案手法は先行研究と比較して正確に分類できているので，提案手
%法は比較的スパースな文脈素性と固有表現の文字種素性との組み合わせを考慮して分
%類が正しくできているのではないかと思われる．
%

The accuracy of \textsc{loc}, however, was lower than that of the log-bilinear
model \cite{primadhanty:2015:acl}. Upon investigating the confusion matrix, we
found that the \textsc{loc} tag was often misclassified as \textsc{per}.
We therefore conclude here that clustering and POS features are necessary to
distinguish these tags.
%そして``LOCATION''の精度があまり高くないが，混同行列を確認したところ，
%``PERSON''タグに間違える事例が多かったので，本実験で使用した素性では組み合わ
%せを考慮しても2 つのタグを区別できなかったのではないかと思われる．
%
%In this experiment, information of named entity candidate's prefix and suffix
%are not include.
%Also \newcite{primadhanty:2015:acl}'s experiment use clustering and POS
%features, although there are dense features than our setting, Factorization
%Machines achieves same degree of accuracy only sparse features.
%固有表現候補の接頭辞や接尾辞の情報も今回の実験では入れていないので，そのよう
%な素性を増やすことでPERSON タグとの混同が減る効果があると期待できる．また，
%\newcite{primadhanty:2015:acl} の実験ではクラスタリング素性や品詞素性を用いて
%いるので，本研究の設定より密な素性が多いが，Factorization Machinesはスパース
%な素性のみでも先行研究と同程度の精度を達成している 

Figure \ref{fig:dimension} plots the F1-score of our proposed method as
dimension $k$ changes for matrix factorization using the same development data
as that of \newcite{primadhanty:2015:acl}.
%図\ref{fig:dimension}はFactorization Machines において，開発データを用いて行
%列分解した次元$k$ を変化させた時の図である．
%
Our method yielded the best F1-score (i.e., 57.1) at $k=5$, whereas
the log-bilinear model achieved the best F1-score (i.e., 61.73) at $k=40$.
These results show that factorization machines require a compact model to
achieve state-of-the-art results on the test set of this corpus.

It would be interesting to point out that the performance of our factorization
machines approach on the development dataset was lower than that of the
log-bilinear model by 4.6 points. This phenomenon may occur because the
log-bilinear model overfits to sparse combinatorial features even with nuclear
norm regularization; further factorization machines typically have better
generalization abilities than those of nuclear norm regularization.
%この図より$k=5$ の場合にF1 スコアが57.1 ポイントと一番高い結果となった．$k=1$
%から$k=5$ までの間は増加し，$k=8$ で一度スコアが下がるが，$k$ の値が大きくな
%るほどF1 スコアも増加している．この結果から，このデータでは$k=5$ の時に行列分
%解した行列が一番良く潜在的な意味を捉えていると言える．
%
%On the other hand, bilinear-model achieve highest accuracy in 40-dimensional,
%Factorization Machines achieves highest accuracy in smaller dimensions than
%10-dimensional.
% \newcite{primadhanty:2015:acl} の対数双線形モデルでは40 次元で一番高い精度を
% 出しており，Factorization Machines では10 次元よりも小さい次元で一番高い精度
% を出している．
%\newcite{primadhanty:2015:acl}'s model can not capture latent meaning from
%matrix decomposition in small dimension, on the other hand Factorization
%Machines can capture. It is advantage that Factorization Machines can achieve
%similar accuracy using smaller parameter.
%これはFactorization Machines では行列分解することで小さい次元で潜在的な意味が
%取れているのに対し，\newcite{primadhanty:2015:acl} のモデルでは40 次元まで潜
%在的な意味が取りきれないのではないかと思われる．Factorization Machines は少な
%いパラメータで行列分解することで同程度の精度を達成することができる，という点
%が優位である．

Both our approach and the methods of \newcite{primadhanty:2015:acl} address the
problem of incorporating sparse combinatorial features by dimension reduction
(i.e., matrix factorization); however, they differ in terms of the objective
function to be optimized.
%\newcite{primadhanty:2015:acl}の核ノルムを用いた正則化モデルとFactorization
%Machines はデータスパースネスの解消のために行列の次元削減をしているという点で
%同じであるが，2 つのモデルでは最適化する目的関数が違っている．
\newcite{primadhanty:2015:acl} use maximum likelihood estimation as an
objective function; whereas other objective functions such as hinge loss can
be used in factorization machines.
%核ノルムを用いた正則化モデルでは対数線形モデルに基づいて，SVD で行列分解して
%いるが，Factorization Machines ではヒンジロスを最適化することで直接的に行列分
%解の最適化ができるという違いが今回の結果を生んだのではないかと思われる．

\section{Conclusion}
In this paper, we proposed the use of factorization machines to handle
the combinations of sparse features in unknown named entity classification.
%この論文では未知の固有表現を認識するタスクにおいて，Factorization Machines を
%用いて行列分解することでスパースな素性同士の組み合わせを考慮した実験を
%行った．
%
Our experimental results showed that we were able to achieve competitive
accuracy to state-of-the-art methods using fewer features and a compact model.
%Factorization Machines を用いることで少ない次元数で先行研究の手法と同程度の精
%度を得られることが分かった．
%
For future work, we aim to extend this framework to sequence labeling, thereby
improving overall named entity recognition.
%Factorization Machines's interaction dimension
%more than $d=2$.
%今後の課題として，\newcite{primadhanty:2015:acl} の実験で使用されていたクラス
%タリング素性や品詞タグを増やして実験することでスパースネスの解消を検証した
%い．

%\section*{Acknowledgments}
%
%Do not number the acknowledgment section.
%This section should not be presented for the submission version.

\clearpage

\bibliography{emnlp2016}
\bibliographystyle{emnlp2016}

\end{document}